\documentclass[runningheads]{llncs}
\usepackage[T1]{fontenc}
\usepackage{graphicx}
\usepackage{booktabs}
\usepackage[misc]{ifsym}

\usepackage{mwe}
\usepackage{listings}

\usepackage{xcolor}
\usepackage{rotating}
\usepackage{tabularx}

\usepackage{longtable}
\usepackage{float}
\usepackage{multirow}
\usepackage{multicol}
\usepackage{colortbl}
\usepackage{makecell}
\usepackage[hidelinks]{hyperref}

\begin{document}

\newcommand{\tvae}{TVAE}
\newcommand{\smote}{SMOTE-NC}
\newcommand{\cart}{CART}
\newcommand{\ctgan}{CTGAN}
\newcommand{\gaussiancopula}{Gaussian Copula}
\newcommand{\gaussiancopulatable}{SDV-GC}

\newcommand{\rf}{\makecell[c]{Random\\Forest}}
\newcommand{\dt}{\makecell[c]{Decission\\Tree}}
\newcommand{\xgb}{XGBoost}
\newcommand{\lgbm}{LightGBM}
\newcommand{\tabfairgan}{Tabfairgan}

\newcommand{\acc}{Accuracy $\uparrow$}
\newcommand{\fone}{F1 $\uparrow$}
\newcommand{\rocauc}{ROC AUC $\uparrow$}
\newcommand{\eqodd}{Equalized Odds $\downarrow$}
\newcommand{\eqoddtable}{Eq. Odds $\downarrow$}
\newcommand{\statpartable}{SP $\downarrow$}
\newcommand{\eqopptable}{Eq. Opp. $\downarrow$}

\newcommand{\classonly}{class}
\newcommand{\classprotected}{class \& protected}
\newcommand{\classprotectedtable}{\makecell[l]{class\\\&\\protected}}

\newcommand{\protectedonly}{protected}
\newcommand{\sameclass}{class (ratio)}

\newcommand{\german}{German credit}
\newcommand{\adult}{Adult}
\newcommand{\dutch}{Dutch census}
\newcommand{\creditdataset}{Credit card clients}

\newcommand{\clf}{CLF}
\newcommand{\samplingmethod}{\makecell[l]{Sampling\\strategy}}
\newcommand{\training}{\makecell[l]{train-set}}
\newcommand{\metrics}{Metrics on test-set (real data)}

% \title{Can Generative AI Mitigate Class Imbalance and Enhance Fairness? A Comparative Study on Tabular Data}

\title{Synthetic Tabular Data Generation for Class Imbalance and Fairness: A Comparative Study}

\titlerunning{Data Generation for Class Imbalance and Fairness}

\author{Emmanouil Panagiotou\inst{1,2}\orcidID{0000-0001-9134-9387} \and Arjun Roy\inst{1,2}\orcidID{0000-0002-4279-9442} \and
Eirini Ntoutsi\inst{2}\orcidID{0000-0001-5729-1003}}
\authorrunning{E. Panagiotou et al.}

\institute{Freie Universität Berlin, Department of Mathematics \& Computer Science, Berlin, Germany \\
\email{emmanouil.panagiotou@fu-berlin.de}\\
\and
Universität der Bundeswehr München, Faculty for Informatik,  Munich, Germany
}

\maketitle              % typeset the header of the contribution

\begin{abstract}
Due to their data-driven nature, Machine Learning (ML) models are susceptible to bias inherited from data, especially in classification problems where class and group imbalances are prevalent. Class imbalance (in the classification target) and group imbalance (in protected attributes like sex or race) can undermine both ML utility and fairness. Although class and group imbalances commonly coincide in real-world tabular datasets, limited methods address this scenario. While most methods use oversampling techniques, like interpolation, to mitigate imbalances, recent advancements in synthetic tabular data generation offer promise but have not been adequately explored for this purpose. To this end, this paper conducts a comparative analysis to address class and group imbalances using state-of-the-art models for synthetic tabular data generation and various sampling strategies. Experimental results on four datasets, demonstrate the effectiveness of generative models for bias mitigation, creating opportunities for further exploration in this direction.

  
\end{abstract}

  % Due to their data-driven nature, Machine Learning (ML) models are susceptible to bias inherited from data, especially in classification problems where class and group imbalances are prevalent. Class imbalance (in the classification target) and group imbalance (in protected attributes like race or sex) can undermine both ML efficacy and fairness. Although class and group imbalances commonly coincide in real-world tabular datasets, existing methods primarily focus on class imbalance, leaving the group imbalance problem unaddressed. While most methods use oversampling techniques, like interpolation, to mitigate class imbalances, they often neglect group imbalances. Recent advancements in synthetic tabular data generation offer promise but have not been explored for this purpose. To this end, this paper conducts a comparative analysis to address class and group imbalances using state-of-the-art models for synthetic tabular data generation and various sampling strategies. Experimental results on four datasets highlight the effectiveness of generative models in mitigating bias, signaling a promising avenue for further research in this area. 


\section{Introduction}
% \manolis{Long submission Length: A full paper of \textbf{10-15 pages} (plus unlimited pages for references). Papers that do not follow the length requirements may be rejected without review}
% \arjun{You can leave comments like this if you want}
% \manolis{I have this paper as a baseline https://doi.org/10.1145/3548785.3548793}

Artificial intelligence (AI) has seamlessly integrated into our daily lives, revolutionizing sectors from personalized online experiences to advanced medical diagnostics. Nonetheless, data collected from real-world sources inherently reflects the biases, prejudices, and inequalities prevalent within society~\cite{tai_survey}. Consequently, ML models trained on such data have the potential to perpetuate and even exacerbate these biases, leading to unfair or discriminatory outcomes~\cite{brackey2019compas}.

One significant data-related challenge that can cause biased predictions is population imbalance. The most common, and easily detectable, is class imbalance, which can lead to poor predictive performance for instances in an under-represented class. Group imbalance, on the other hand, might not directly affect the utility of a model in terms of overall accuracy, but it can lead to unfair treatment of minority groups characterized by some protected attribute (e.g. sex or race). Several methods have been proposed to address both class-imbalanced learning \cite{johnson2019survey}, and fairness \cite{mehrabi2021survey}, yet only a few works study their overlap, which is very common for real-world tabular datasets \cite{tai_survey}. Most of these methods are model-specific, meaning they change the workings of existing models to increase performance in minority and majority groups.

Despite the prevalent use of generative methods for synthesizing tabular data, there remains a gap in evaluating their influence on group fairness and class imbalance. In this work, we perform a comparative analysis of \emph{model-independent} generative techniques using oversampling to address class and group imbalances in tabular datasets. While most existing methods rely on the Synthetic Minority Oversampling Technique (SMOTE) \cite{smote} for generating additional samples, more recent works on generative AI have developed numerous alternatives for synthesizing tabular data \cite{endres2022synthetic}. We cover five generative methods: the probabilistic Gaussian Copula \gaussiancopulatable\ \cite{patki2016synthetic}, two deep learning models \ctgan\ and \tvae\ \cite{tvae} based on GANs and VAEs respectively, generative non-parametric classification and regression trees \cart\ \cite{cart}, and the conventional \smote\ \cite{smote} for oversampling via interpolation. We also define four sampling strategies for these generative methods and evaluate their performance on ML utility and fairness using four real-world tabular datasets. Our results are benchmarked against training on the original (real) data and a state-of-the-art fair data generator, \tabfairgan\ \cite{tabfairgan}. We conclude with an experiment on intersectional fairness, examining the scenario where multiple protected attributes coexist. The full code for this study is available under,  \href{https://github.com/Panagiotou/FairAugment}{github.com/Panagiotou/FairAugment}.

The rest of this paper is organized as follows. We describe all relevant works related to fairness, class imbalance, and generative methods in Section~\ref{sec:related}, we present the problem formulation, dataset details, and evaluation metrics in Section~\ref{sec:background}, we define all sampling strategies in Section~\ref{sec:method}, and include all experiments and results in Section~\ref{sec:experiments}. We conclude the paper with a discussion and opportunities for future work in Section~\ref{sec:discussion}.

\section{Related Work}
\label{sec:related}

Due to the data-driven nature of ML, inherent biases within the data frequently get amplified or perpetuated, resulting in unfair decision-making. Such bias can arise from imbalanced populations regarding the target class labels, or subgroups defined by protected attributes (e.g. sex, race, etc.). This has led to new research directions towards mitigating such bias and developing fairness-aware models \cite{mehrabi2021survey,bellamy2019ai}. In this section, we cover all related work focusing on overcoming class imbalance, group imbalance (fairness-aware ML), and their simultaneous occurrence. Additionally, we cover synthetic tabular data generation methods, that are relevant to our comparative study. 

\subsection{Fairness-aware ML}
In our study, we focus on \emph{group fairness}, which considers parity over different groups of individuals, distinguished by one or more protected attributes, such as sex, race, age, etc. Several metrics have been defined to measure the group fairness of a classification model (see Section~\ref{sec:metrics}). While there are various methods for enhancing group fairness, the main focus is on i) creating methods that specifically optimize for fairness \cite{iosifidis2019adafair}, by incorporating constraints to existing models, for example via adding fairness objectives to the loss function \cite{roy2022learning}, and ii) model-agnostic, pre-processing methods \cite{dwork2012fairness,kamiran2012data,sonoda2023fair} that overcome bias by modifying the training data. 

We focus on the second approach, specifically generative pre-processing methods, which rely solely on the training data and mitigate bias by augmenting the existing training data with new samples or sampling an entirely new dataset. For example, the GAN-based \tabfairgan\ \cite{tabfairgan} optimizes for accuracy and fairness with consecutive training phases. Once fitted, an entire synthetic dataset is sampled. However, while all of these methods address group fairness, they do not take class imbalance into account.

\subsection{Class imbalance in ML} Class imbalance is a common problem in classification problems \cite{tai_survey}, where a large percentage of the data belongs to a specific class. This scenario is encountered in various domains, such as clinical studies, where the minority class (indicating illness) is under-represented, compared to the majority class (representing healthy individuals). To tackle this issue, similar to fairness methods, many approaches resort to pre-processing techniques like over/under-sampling to mitigate the bias towards the majority class \cite{lemaavztre2017imbalanced,he2008adasyn}.

In general, under-sampling methods are not typically favored due to the potential loss of crucial information, which can degrade performance. Similarly, naive over-sampling techniques, such as simply duplicating individuals in the minority class, may lead to overfitting. To overcome this problem, the renowned Synthetic Minority Oversampling Technique (SMOTE) was proposed \cite{smote}. SMOTE operates by interpolating between random instances in the minority class and their K-nearest neighbors. This concept has led to various extensions \cite{fernandez2018smote} which for example sample specific regions, such as those close to the decision boundary \cite{he2008adasyn}, or more sparse areas of the feature space \cite{douzas2018improving}. While such methods improve ML utility by reducing bias towards a certain class, they do not account for group fairness.

\subsection{Fairness and class imbalance in ML}
\label{sec:fair_b_ml}
Bias in the data related to fairness and class imbalance are not mutually exclusive. More often than not, they occur simultaneously \cite{tai_survey}, leading to extreme population imbalance for individuals from a minority group who are assigned underrepresented class labels. For example, in the popular \adult\ dataset, females with a high-income class label are the most under-represented subgroup, accounting for only $11\%$ of the total data (see Figure~\ref{fig:sampling_strategies}). These populations can become even smaller under "intersectional-fairness" when more than one protected attribute exists \cite{roy2022multi} or for multi-class classification.

To address this issue, various fair class-balancing methods like FSMOTE \cite{fsmote}, FAWOS \cite{salazar2021fawos}, and other extensions \cite{sonoda2023fair,yan2020fair}, have been proposed. The goal is to overcome both fairness and class imbalance via model-independent oversampling. Yet, most of these methods either employ the SMOTE interpolation technique for oversampling or assume a common (discrete) feature type \cite{xu2023ffpdg}, rendering them unsuitable for handling numerical or mixed feature spaces, which are very common in tabular datasets \cite{tai_survey}. 

Nonetheless, recently several generative models have been proposed for generating synthetic mixed tabular data, for example, based on neural networks \cite{tvae}, classification trees \cite{cart}, probabilistic approaches \cite{mdgmm}, or even large language models \cite{llm}. In this work, we evaluate such synthetic tabular data generation methods (defined in Section~\ref{sec:generative_methods}) for class imbalance and fairness, while considering different sampling strategies.

% \subsection{Fairness overall}
% % https://www.sciencedirect.com/science/article/pii/S0020025523006448#bbr0030
% \subsection{Class imbalance overall}

% \subsection{Fair class balancing techniques}

\subsection{Synthetic Tabular Data Generation Methods}
\label{sec:generative_methods}

Various methods have been proposed to learn to generate tabular data \cite{endres2022synthetic}. Compared to other modalities such as images or text, tabular datasets consist of a mixture of discrete and continuous feature types, which are difficult to model. Our analysis covers recent approaches for efficient and effective tabular data generation, encompassing state-of-the-art parametric and non-parametric methods.

\begin{itemize}
    \item \textbf{\gaussiancopulatable:} Various continuous distributions (e.g. uniform, exponential, etc.) are considered to model all features (discrete features are not explicitly handled, but transformed into continuous). Subsequently, a multivariate Gaussian Copula is used to estimate the covariance between all features. The covariance matrix and the feature distributions are used to sample new synthetic data \cite{patki2016synthetic}.
    
    \item \textbf{\ctgan:} The typical generator/critic neural network architecture for Generative Adversarial Networks (GANs) is adapted to learn to generate tabular data. Mode-specific normalization is used during training to overcome imbalances and avoid mode collapse \cite{tvae}.
    
    \item \textbf{\tvae:} The Tabular Variational Autoencoder \cite{tvae} trains an encoder/decoder neural network to learn a low-dimensional Gaussian latent space, which is used for sampling new instances through the trained decoder. 
    
    \item \textbf{\cart:} A Classification and Regression Tree \cite{cart} method for consecutive column-wise data generation via sampling in the leaves, especially suitable for learning inter-dependencies between mixed data due to its non-parametric nature.
    
    \item \textbf{\smote:} SMOTE (Synthetic Minority Over-sampling Technique) \cite{smote} is a non-parametric method that generates new samples by interpolating between line segments connecting real instances. The same paper introduces the \smote\ variant, which can support mixed (but not solely discrete) feature spaces. 
\end{itemize}

\section{Background}
\label{sec:background}
We assume a tabular dataset $T$ containing $N_c$ continuous columns $\{c_1, c_2, \ldots, c_{N_c}\}$ and $N_d$ discrete columns $\{d_1, d_2, \ldots, d_{N_d-1}, d_{prot}\}$ (including categorical, binary, and ordinal features). Additionally, we assume one binary \emph{protected attribute} $d_{prot} \in \{0,1\}$ (e.g. the sex of an individual), and a binary class label $Y \in \{0,1\}$. Given such a dataset, any given ML classifier $f()$ can be trained in a supervised manner, on input-target pairs $x_j = \{c_1, c_2, \ldots, c_{N_c}, d_1, d_2, \ldots, d_{N_d-1}\}$ and $y_j \in \{0,1\}$, $j=\{1, 2, \ldots, n\}$ (the protected attribute is not used during training). Since the class label and the protected attribute are binary features, they partition the tabular dataset $T$ into $|d_{prot} \times Y| = 4$ subgroups $[T_{\scriptscriptstyle 00}, T_{\scriptscriptstyle 01}, T_{\scriptscriptstyle 10}, T_{\scriptscriptstyle 11}]$. 

A generative model $G$ fitted on some subset $\tilde{T}$ of the dataset $T$, can sample $\tilde{n}$ synthetic rows that comprise a synthetic dataset $\tilde{T}_{syn} = G(\tilde{T}, \tilde{n})$. Further, we refer to a \emph{sampling strategy} $S(n_{\scriptscriptstyle 00}, n_{\scriptscriptstyle 01}, n_{\scriptscriptstyle 10}, n_{\scriptscriptstyle 11})$ as the method that dictates the number of synthetic samples to be generated from each subgroup in $T$, to generate a synthetic dataset $T_{syn} = [G(T_{\scriptscriptstyle 00},n_{\scriptscriptstyle 00}), G(T_{\scriptscriptstyle 01}, n_{\scriptscriptstyle 01}), G(T_{\scriptscriptstyle 10}, n_{\scriptscriptstyle 10}), G(T_{\scriptscriptstyle 11}, n_{\scriptscriptstyle 11})]$. The objective of a sampling strategy in our case, is to create an \emph{augmented} final training dataset, denoted as $T_{aug} = T \cup T_{syn}$ which aims to enhance the classifier's performance regarding class imbalance and fairness. We refer to the proportion of the synthetic samples in the augmented dataset as the \emph{augmentation ratio} $r_{aug} = |T_{syn}|/|T_{aug}|$.

In our study, we define and compare various over-sampling methods (Section~\ref{sec:method}) dictated by the generative models and sampling strategies $G, S$, aiming to correct both class and group imbalance.

\subsection{Datasets}
We use four real-world tabular datasets, frequently used in fairness-aware learning \cite{tai_survey}. These datasets comprise demographic attributes of individuals, aimed at predicting their financial status, such as occupation, income, credit score, etc. In Table~\ref{tab:datasets} we list the basic characteristics of all datasets, namely, the \emph{\adult}, \emph{\german}, \emph{\dutch}, and \emph{\creditdataset}. The protected attribute chosen for all datasets is the binary feature "sex" (Male/Female). We observe class imbalance for the \emph{\adult}, \emph{\german}, and \emph{\creditdataset}\ datasets, as well as, a mixed feature space. The \emph{Dutch census} dataset exhibits a less pronounced class imbalance and includes solely discrete features. Both class and group imbalances for all datasets are visualized in the first column of Fig~\ref{fig:sampling_strategies}.

\begin{table}[h]
\centering
\begin{tabular}{|l|c|c|c|c|c|c|c|}
\hline
\textbf{Dataset} & \textbf{\#Instances} & \makecell[c]{\textbf{\#Attributes}\\$\mathbf{N_d/N_c}$} & \makecell[c]{\textbf{Class Ratio}\\\textbf{(+)}} & \makecell[c]{\textbf{Protected}\\\textbf{(Attribute)}} & \textbf{Target Class} \\ \hline
\adult & 45k & 9/6  & 1:3.03 & sex & Income \\ \hline
\german & 1k & 14/7  & 2.33:1 & sex & Credit score \\ \hline
\dutch & 60k & 12/0  & 1:1.10 & sex & Occupation \\ \hline
\makecell[l]{Credit card\\clients} & 30k & 10/14  & 1:3.52 & sex & \makecell[c]{Default\\payment} \\ \hline
\end{tabular}
\caption{Overview of all real datasets used in our comparative study}
\label{tab:datasets}
\end{table}

\subsection{Evaluation Metrics}
\label{sec:metrics}
To evaluate the quality of the synthetic data regarding ML utility and fairness, we measure the performance of ML models on the downstream binary classification task for each dataset. In terms of utility, we measure the \emph{Accuracy} and \emph{ROC AUC} score. The last is more suitable for evaluation under class imbalance, as it takes true/false-positive/negative rates into account. With respect to group fairness, we employ widely used fairness metrics, namely equalized odds \cite{hardt2016equality} (Eq. Odds), statistical parity \cite{dwork2012fairness} (SP), and equal opportunity \cite{hardt2016equality} (Eq. Opp.). We define all fairness metrics below:

\begin{itemize}
    \item \underline{Equalized Odds (Eq. Odds):}\\
    Assesses the difference between true positive rates (for positive class) and false positive rates (for negative class) for different groups. \\
    Eq. Odds = $|P[f(X)=1|Y=1, d_{prot}=0] - P[f(X)=1|Y=1, d_{prot}=1]| + \\ |P[f(X)=1|Y=0, d_{prot}=0] -  P[f(X)=1|Y=0, d_{prot}=1]|$
    \item \underline{Statistical Parity (SP):}\\
    Measures whether the probability of a favorable outcome is consistent across different groups defined by the sensitive attribute.\\
    SP = $P[f(X)=1| d_{prot}=0] - P[f(X)=1| d_{prot}=1]|$
    \item \underline{Equal Opportunity (Eq. Opp.):}\\
    Measures the difference between the true positive rates (sensitivity) across sensitive attribute groups.\\
    Eq. Opp. = $|P[f(X)=1|Y=1, d_{prot}=0] - P[f(X)=1|Y=1, d_{prot}=1]|$

\end{itemize}

All utility metrics should be maximized (the closer to $1$ the better), while fairness metrics are minimized (the closer to $0$ the better), as they measure differences in performance between subgroups. Since we compare model-independent generative methods, any downstream classification model can be used for evaluation. We choose XGBoost \cite{chen2016xgboost}, a state-of-the-art gradient boosting model for tabular data classification.

\section{Sampling strategies}
\label{sec:method}
\begin{figure}[h]
  \centering
  \includegraphics[width=0.9\linewidth]{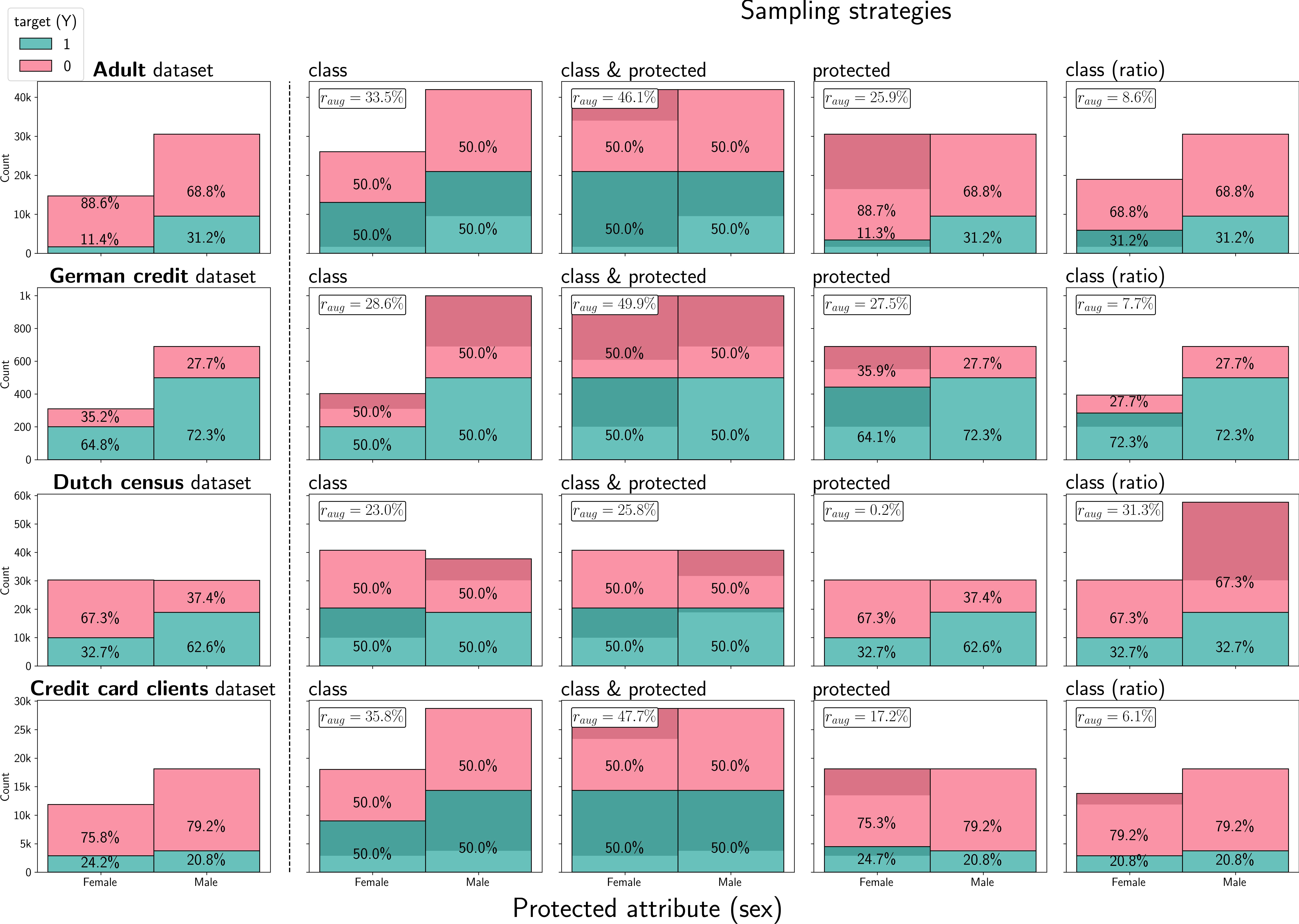}
  \caption{Distributions of class and group imbalance for each real dataset (first column) along with final augmented dataset for each sampling strategy.}
  \label{fig:sampling_strategies}
\end{figure}

All generative methods covered in our study (described in Section~\ref{sec:generative_methods}) can be trained on a set of tabular data, and then used to generate an arbitrary number of synthetic samples. Given our assumption of a single binary protected attribute and a binary classification task, this results in $4$ homogeneous subgroups for each dataset. Additionally, as defined in Section~\ref{sec:background}, a sampling strategy dictates the number of synthetic samples to draw from each subgroup, to create the final augmented training set. In this work, we propose four such sampling strategies aimed at addressing class imbalance, group imbalance, or both. Namely, \emph{\classonly} and \emph{\protectedonly}, sample data to achieve class, and group balance, respectively. Furthermore, \emph{\classprotected}, and \emph{\sameclass}, sample synthetic data to achieve both class and group parity. We define each sampling strategy in detail hereafter and provide a visual representation of the final distributions of the augmented data for each sampling strategy on all datasets in Figure~\ref{fig:sampling_strategies}. 

\begin{itemize}
    \item \textbf{\classonly:} Separately for each group (Male/Female) we sample instances for the minority class, to match the number of instances in the majority class. Therefore, we achieve a $50/50$ class balance for each group.   

    \item \textbf{\classprotected:} 
    For the largest group (e.g. Male) we sample instances for the minority class, to match the number of instances in the majority class. For all other groups (e.g. Female) we sample for both the majority class and the minority class, to match the number of instances in the majority class in the largest group. Therefore, we achieve the same number of samples for all $4$ subgroups. It is worth noting that the \emph{\classprotected} strategy is described and used by the FSMOTE method \cite{fsmote} (refer to Section~\ref{sec:fair_b_ml}). 
    
    \item \textbf{\protectedonly:} 
    We do not sample for the largest group (e.g. Male), but only for all other groups (e.g. Female), to match the number of instances in the largest group, without considering class labels. Therefore, we achieve the same number of instances for all groups. 
    
    \item \textbf{\sameclass:}
    We do not sample for the largest group (e.g. Male), but only for all other groups (e.g. Female), to match the class ratio of the largest group.  
    Therefore, we achieve that all groups have the same class ratio as the largest group in the original dataset. 

\end{itemize}

For each sampling strategy in the figure, we report the \emph{augmentation ratio} $r_{aug}$, as defined in Section~\ref{sec:background}, i.e. the percentage of synthetic samples in the final augmented dataset. To visualize this, the number of synthetic samples in each bar plot is depicted with a darker color than the real data.

\section{Experiments and results}
\label{sec:experiments}
% \manolis{with \textbf{bold} I have best, with \underline{underline} second best, with blue color I have settings performing better than training on real data}

% \manolis{tables too wide}
% \manolis{Here I will add an experiment on runtime for train/sample for generative methods}

% ['gaussian_copula', 'ctgan', 'tvae', 'cart', 'smote']
% train mean [2.235555648803711, 165.434029229482, 63.34298477172852, 1.02997891108195, 0.014052232106526693]
% train std [0.13489924717475418, 1.7717362477805154, 2.590806199151657, 0.021446196435492268, 0.00025537084631051097]
% sample mean [0.1486209233601888, 0.16410729090372722, 0.08441376686096191, 0.35485196113586426, 17.315525658925374]
% sample std [0.005069463502613001, 0.04722246498290099, 0.009870092336613944, 0.006311444243275272, 0.31801760417289193]

In this section, we present our comparative study, evaluating all generative methods and sampling strategies under utility and fairness. Additionally, we perform an experiment on intersectional fairness, taking multiple protected attributes into account. We conclude with a runtime comparison of all generative methods.

\subsection{Experimental setup}

We perform experiments for all four datasets, four sampling methods, and five generative models. To ensure robustness, each experiment on the downstream task is 3-fold cross-validated and repeated two times over different random seeds. We report average results over all repetitions, highlighting the best results in bold, and underlining the second-best. For the accumulated results of Section~\ref{sec:all_res}, we further shade with \emph{blue color} the experiments on synthetic data, which exhibit better performance than training on the original real data (first row). 

All experiments are conducted on a single machine equipped with a 12th Gen Intel(R) Core(TM) i9 processor and a Nvidia GeForce RTX 3080 Ti GPU.

\subsection{Accumulated results on all datasets}
\label{sec:all_res} 

The following Table~\ref{tab:results_p1} and Table~\ref{tab:results_p2} show the results of our comparative study for the \emph{\adult}\ and \emph{\german}\ datasets, and the \emph{\dutch}\ and \emph{\creditdataset}\ datasets, respectively. As previously mentioned, we present average metrics for all sampling strategies and generative methods. The first two rows in each table (for each dataset) correspond to baselines, i.e. training the classifier using the real data, and synthetic data generated with \tabfairgan\ \cite{tabfairgan}. Subsequent rows display results for augmented training data generated through various combinations of the five generative methods and four sampling strategies. We highlight in blue the experiments with superior performance compared to training on the real dataset. Testing (evaluation) is always performed on, previously-unseen, test data from the real dataset.

We interpret the results based on the following criteria:

\noindent{\textbf{Accuracy:}} An initial observation of the results suggests an overall decrease in classifier \emph{accuracy} across datasets when using synthetic data. This is substantiated by relevant literature \cite{tvae}, and can be ascribed to the introduction of out-of-distribution synthetic data by the generative methods.

\noindent{\textbf{ROC AUC (class imbalance):}} Sampling strategies focusing on \emph{class balancing}, such as \emph{\classonly}\ and \emph{\classprotected}\, improve the ROC AUC score for imbalanced datasets. For the \emph{dutch}\ dataset, we do not observe any improvement, due to the lack of inherent class imbalance in the data (see Table~\ref{tab:datasets}). Notably, the best ROC AUC score in most cases is achieved with \cart-generated data.  

\noindent{\textbf{Fairness:}} We observe that the \tabfairgan\ baseline, although specifically optimized for statistical parity (SP), can also lead to improvements in terms of Eq. Odds and Eq. Opp. However, it is evident that using generative methods, especially with \sameclass\ sampling, leads to superior fairness metrics while maintaining higher utility (ROC AUC). This can be attributed to the fact that for most datasets (excluding \dutch), fewer synthetic samples are needed to achieve equal class ratios between different subgroups, i.e. a lower $r_{aug}$ (see Fig.~\ref{fig:sampling_strategies}). On the other hand, the \classprotected\ strategy requires the highest number of synthetic samples to maintain class and group balance. This increases the risk of producing out-of-distribution samples, which can degrade performance.

\noindent{\textbf{Generative methods:}} The non-parametric \cart\ model emerges as the top performer in most cases. Notably, despite its simplicity, \smote\ demonstrates performance similar to deep methods, i.e. \tvae, \ctgan. However, it is not applicable for datasets with exclusively discrete feature spaces, such as the \dutch\ dataset.

To summarize, sampling strategies like \emph{\classonly} and \emph{\classprotected} improve ROC AUC for imbalanced datasets, and \cart\ often achieves the best results. The \sameclass\ strategy enhances fairness metrics, generating fewer synthetic samples and maintaining utility.

% Nonetheless, it is evident that, as expected, the \emph{\classprotected} sampling strategy is the best trade-off solution between ML efficacy and fairness. 

% Nonetheless, the \emph{\sameclass} sampling strategy consistently demonstrates superior performance in fairness metrics compared to others in most cases. This advantage is likely attributed to the fact that, in general, fewer synthetic samples are required for this strategy in contrast to \emph{\classprotected}. For instance, in the \adult\ dataset, only approximately $5.000$ instances of females with the class label $Y=1$ are generated, whereas the \classprotected\ strategy necessitates generating $35.000$ synthetic samples overall (refer to Figure~\ref{fig:sampling_strategies}). The reduced requirement for synthetic samples reduces uncertainty stemming from the generative method. Notably, the \sameclass strategy on the adult dataset achieves up to approximately, $35\%$ improvement in equalized odds, $20\%$ improvement in statistical parity, and $60\%$ improvement in equal opportunity.

\subsection{Intersectional fairness}

\begin{figure}[h]
  \centering
  \includegraphics[width=0.48\linewidth]{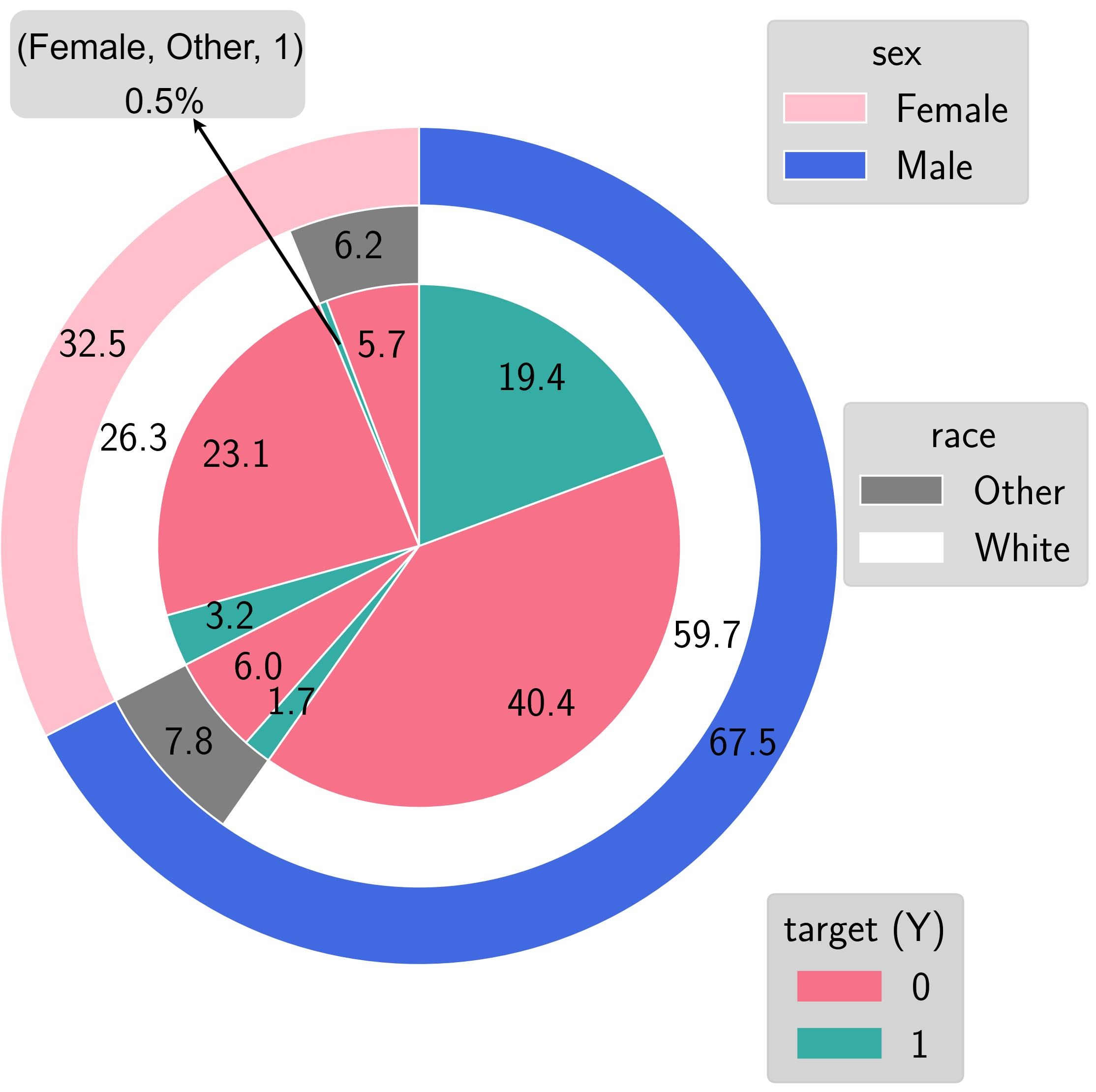}
  \caption{Sex, race, and class subgroup percentage distributions of the adult dataset.}
  \label{fig:adult_distr}
\end{figure}

In all previous experiments, we assume a single binary protected attribute and class label, leading to $4$ subgroups (see Section~\ref{sec:background}). Nonetheless, in some cases multiple protected attributes can exist, partitioning the data into further groups. To study this effect, we conduct an experiment on the \adult\ dataset, with race = \{White, Other\} as the additional protected attribute, splitting the data into $8$ subgroups. In Table~\ref{tab:results_intersectional}, we present the results when using \emph{race} and \emph{sex \& race} (intersection) as protected attributes. In line with our previous results, we observe that the \sameclass\ strategy and \cart\ generative method significantly enhance fairness without compromising utility. The non-parametric nature of \cart\ enables consistent generation even from the most under-represented subgroups under extreme data scarcity. For example, in the \adult\ dataset, the subgroup (sex, race, class) = (Female, Other, 1) accounts for only $0.5\%$ of the total data, i.e. under $200$ instances, as seen in Fig.~\ref{fig:adult_distr}.

\subsection{Runtime comparison}
We conclude our experiments by performing a runtime comparison of all generative methods. We report runtime for, i) training (fitting) on the \adult\ dataset, ii) sampling $10.000$ synthetic instances, and iii) training and sampling, since this overall runtime is the most significant metric in our comparison. From Table~\ref{tab:runtime} it becomes evident that the \cart\ method outperforms all others significantly in terms of overall runtime.

% ['gaussian_copula', 'ctgan', 'tvae', 'cart', 'smote']
% [2.2590940316518147, 215.49478891690572, 73.52785964012146, 1.0198370774586996, 0.016304747263590495]
% [0.137682621283486, 49.118277427023344, 7.855912847482573, 0.01790332718083524, 0.004422130339486343]
% [0.16322725613911945, 0.15528831481933594, 0.15989872614542644, 0.32647811571756996, 18.955435276031494]
% [0.01396410154343295, 0.027763671747503042, 0.27184605944230217, 0.0020027551568974223, 1.6070019652443603]
% [2.4223212877909344, 215.65007723172505, 73.68775836626689, 1.3463151931762696, 18.971740023295084]
% [0.13851753560966126, 49.132941493935654, 7.957795919995051, 0.018595690990526163, 1.6092161336178314]

\begin{table}[ht!]
\caption{Training and sampling runtime comparison for all generative methods on the \adult\ dataset.}
\label{tab:runtime}
\centering
\begin{tabular}{l l l l}
\hline
Model & Training time $\downarrow$ & Sampling time $\downarrow$ & Overall time $\downarrow$\\
\hline
\tabfairgan & 187.521 \tiny{$\pm$ 7.64} & \textbf{0.011} \tiny{$\pm$ 0.005} & 187.533 
\tiny{$\pm$ 7.64}\\
\hline
\gaussiancopulatable & 2.259 \tiny{$\pm$ 0.137} & 0.163 \tiny{$\pm$ 0.013} & \underline{2.422} \tiny{$\pm$ 0.138}\\
\ctgan & 215.49 \tiny{$\pm$ 49.118} & \underline{0.155} \tiny{$\pm$ 0.027} & 215.65 \tiny{$\pm$ 49.132}\\
\tvae & 73.527 \tiny{$\pm$ 7.855} & 0.159 \tiny{$\pm$ 0.271} & 73.687 \tiny{$\pm$ 7.957}\\
\cart & \underline{1.019} \tiny{$\pm$ 0.017} & 0.326 \tiny{$\pm$ 0.002} & \textbf{1.346} \tiny{$\pm$ 0.018}\\
\smote & \textbf{0.016} \tiny{$\pm$ 0.004} & 18.955 \tiny{$\pm$ 1.607} & 18.971 \tiny{$\pm$ 1.609}\\
\hline
\end{tabular}
\end{table}

\begin{table*}[ht!]
\caption{Results for \adult\ and \german\ datasets.}
\label{tab:results_p1}
\begin{tabular}{l l c c c c c}
\hline
\multicolumn{7}{c}{\cellcolor{black!15}\adult\ dataset}\\
\samplingmethod & \training & \multicolumn{5}{c}{\metrics} \\
&  & \multicolumn{5}{c}{sex} \\
& & \acc & \rocauc & \eqoddtable & \statpartable & \eqopptable \\
\hline & \multirow{1}{*}{Real} & \textbf{0.868} & 0.798 & 0.122 & 0.178 & 0.059 \\
\hline
 & \multirow{1}{*}{\tabfairgan} & 0.539 & 0.626 & \cellcolor{blue!15}0.119 & \cellcolor{blue!15}\underline{0.123} & \cellcolor{blue!15}\textbf{0.029} \\
\hline
\hline
\multirow{5}{*}{\classonly} & \multirow{1}{*}{\gaussiancopulatable} & 0.855 & 0.767 & \cellcolor{blue!15}0.116 & \cellcolor{blue!15}0.149 & 0.066 \\
 & \multirow{1}{*}{\ctgan} & 0.842 & 0.791 & 0.147 & 0.189 & 0.068 \\
 & \multirow{1}{*}{\tvae} & 0.846 & 0.768 & 0.131 & \cellcolor{blue!15}0.168 & 0.063 \\
 & \multirow{1}{*}{\cart} & 0.836 & \cellcolor{blue!15}\underline{0.804} & 0.155 & \cellcolor{blue!15}0.174 & 0.076 \\
 & \multirow{1}{*}{\smote} & 0.839 & 0.788 & 0.133 & 0.181 & \cellcolor{blue!15}0.052 \\
\hline
\multirow{5}{*}{\classprotectedtable} & \multirow{1}{*}{\gaussiancopulatable} & 0.854 & 0.766 & 0.122 & \cellcolor{blue!15}0.161 & 0.065 \\
 & \multirow{1}{*}{\ctgan} & 0.853 & 0.792 & 0.161 & 0.191 & 0.083 \\
 & \multirow{1}{*}{\tvae} & 0.844 & 0.772 & 0.169 & 0.188 & 0.082 \\
 & \multirow{1}{*}{\cart} & 0.836 & \cellcolor{blue!15}\textbf{0.814} & 0.138 & 0.194 & \cellcolor{blue!15}0.048 \\
 & \multirow{1}{*}{\smote} & 0.840 & 0.790 & 0.127 & 0.188 & \cellcolor{blue!15}0.043 \\
\hline
\multirow{5}{*}{\protectedonly} & \multirow{1}{*}{\gaussiancopulatable} & 0.853 & 0.754 & 0.140 & \cellcolor{blue!15}0.155 & 0.089 \\
 & \multirow{1}{*}{\ctgan} & 0.854 & 0.755 & 0.122 & \cellcolor{blue!15}0.148 & 0.072 \\
 & \multirow{1}{*}{\tvae} & 0.859 & 0.768 & \cellcolor{blue!15}0.116 & \cellcolor{blue!15}0.157 & 0.064 \\
 & \multirow{1}{*}{\cart} & 0.856 & 0.763 & \cellcolor{blue!15}0.116 & \cellcolor{blue!15}0.154 & 0.065 \\
 & \multirow{1}{*}{\smote} & 0.857 & 0.758 & 0.160 & \cellcolor{blue!15}0.162 & 0.105 \\
\hline
\multirow{5}{*}{\sameclass} & \multirow{1}{*}{\gaussiancopulatable} & 0.857 & 0.766 & \cellcolor{blue!15}0.102 & \cellcolor{blue!15}0.151 & \cellcolor{blue!15}0.051 \\
 & \multirow{1}{*}{\ctgan} & \underline{0.860} & 0.776 & \cellcolor{blue!15}0.103 & \cellcolor{blue!15}0.152 & \cellcolor{blue!15}0.055 \\
 & \multirow{1}{*}{\tvae} & 0.857 & 0.767 & \cellcolor{blue!15}\textbf{0.079} & \cellcolor{blue!15}0.142 & \cellcolor{blue!15}\underline{0.033} \\
 & \multirow{1}{*}{\cart} & 0.857 & 0.780 & 0.137 & \cellcolor{blue!15}\textbf{0.117} & 0.109 \\
 & \multirow{1}{*}{\smote} & 0.854 & 0.768 & \cellcolor{blue!15}\underline{0.095} & \cellcolor{blue!15}0.124 & 0.062 \\

\hline
\multicolumn{7}{c}{\cellcolor{black!15}\german\ dataset}\\
\hline

\samplingmethod & \training & \multicolumn{5}{c}{\metrics} \\
&  & \multicolumn{5}{c}{sex} \\
& & \acc & \rocauc & \eqoddtable & \statpartable & \eqopptable \\
\hline & \multirow{1}{*}{Real} & 0.753 & 0.677 & 0.088 & 0.058 & 0.051 \\
\hline
 & \multirow{1}{*}{\tabfairgan} & 0.639 & 0.501 & 0.090 & \cellcolor{blue!15}0.042 & 0.056 \\
\hline
\hline
\multirow{5}{*}{\classonly} & \multirow{1}{*}{\gaussiancopulatable} & 0.731 & \cellcolor{blue!15}0.686 & 0.097 & \cellcolor{blue!15}\textbf{0.019} & \cellcolor{blue!15}0.045 \\
 & \multirow{1}{*}{\ctgan} & 0.732 & \cellcolor{blue!15}\textbf{0.704} & 0.102 & \cellcolor{blue!15}0.037 & \cellcolor{blue!15}0.041 \\
 & \multirow{1}{*}{\tvae} & 0.736 & 0.675 & 0.098 & 0.067 & \cellcolor{blue!15}0.044 \\
 & \multirow{1}{*}{\cart} & 0.730 & \cellcolor{blue!15}\underline{0.695} & 0.127 & \cellcolor{blue!15}0.057 & \cellcolor{blue!15}0.047 \\
 & \multirow{1}{*}{\smote} & 0.736 & \cellcolor{blue!15}0.687 & \cellcolor{blue!15}\textbf{0.073} & \cellcolor{blue!15}0.038 & \cellcolor{blue!15}0.041 \\
\hline
\multirow{5}{*}{\classprotectedtable} & \multirow{1}{*}{\gaussiancopulatable} & 0.726 & \cellcolor{blue!15}0.683 & 0.103 & \cellcolor{blue!15}\underline{0.024} & \cellcolor{blue!15}\underline{0.040} \\
 & \multirow{1}{*}{\ctgan} & 0.730 & \cellcolor{blue!15}0.687 & 0.099 & \cellcolor{blue!15}0.052 & \cellcolor{blue!15}0.045 \\
 & \multirow{1}{*}{\tvae} & 0.749 & \cellcolor{blue!15}0.690 & 0.113 & 0.065 & 0.052 \\
 & \multirow{1}{*}{\cart} & 0.722 & \cellcolor{blue!15}0.680 & 0.104 & \cellcolor{blue!15}0.045 & \cellcolor{blue!15}0.045 \\
 & \multirow{1}{*}{\smote} & 0.738 & \cellcolor{blue!15}0.681 & 0.121 & \cellcolor{blue!15}0.033 & 0.063 \\
\hline
\multirow{5}{*}{\protectedonly} & \multirow{1}{*}{\gaussiancopulatable} & 0.733 & 0.668 & 0.107 & \cellcolor{blue!15}0.054 & 0.063 \\
 & \multirow{1}{*}{\ctgan} & 0.733 & 0.652 & 0.120 & 0.062 & 0.053 \\
 & \multirow{1}{*}{\tvae} & 0.743 & 0.666 & 0.113 & 0.060 & \cellcolor{blue!15}0.050 \\
 & \multirow{1}{*}{\cart} & 0.737 & 0.664 & 0.132 & 0.064 & \cellcolor{blue!15}\textbf{0.033} \\
 & \multirow{1}{*}{\smote} & 0.739 & 0.669 & 0.108 & \cellcolor{blue!15}0.046 & 0.053 \\
\hline
\multirow{5}{*}{\sameclass} & \multirow{1}{*}{\gaussiancopulatable} & \cellcolor{blue!15}\underline{0.760} & 0.676 & \cellcolor{blue!15}\underline{0.084} & 0.061 & \cellcolor{blue!15}0.048 \\
 & \multirow{1}{*}{\ctgan} & \cellcolor{blue!15}\textbf{0.763} & \cellcolor{blue!15}0.683 & 0.097 & \cellcolor{blue!15}0.044 & \cellcolor{blue!15}0.045 \\
 & \multirow{1}{*}{\tvae} & 0.748 & 0.675 & 0.118 & \cellcolor{blue!15}0.037 & 0.053 \\
 & \multirow{1}{*}{\cart} & 0.753 & 0.675 & 0.104 & 0.060 & 0.056 \\
 & \multirow{1}{*}{\smote} & 0.749 & 0.667 & 0.112 & \cellcolor{blue!15}0.048 & 0.055 \\
\hline

\end{tabular}
\end{table*}

\begin{table*}[ht!]
\caption{Results for \dutch\ and \creditdataset\ datasets.}
\label{tab:results_p2}
\centering
\begin{tabular}{l l c c c c c}
\hline
\multicolumn{7}{c}{\cellcolor{black!15}\dutch\ dataset}\\

\hline
\samplingmethod & \training & \multicolumn{5}{c}{\metrics} \\
&  & \multicolumn{5}{c}{sex} \\
& & \acc & \rocauc & \eqoddtable & \statpartable & \eqopptable \\
\hline & \multirow{1}{*}{Real} & \textbf{0.819} & \textbf{0.817} & 0.092 & 0.189 & 0.049 \\
\hline
 & \multirow{1}{*}{\tabfairgan} & 0.804 & 0.801 & \cellcolor{blue!15}\textbf{0.078} & \cellcolor{blue!15}0.184 & \cellcolor{blue!15}\textbf{0.037} \\
\hline
\hline
\multirow{4}{*}{\classonly} & \multirow{1}{*}{\gaussiancopulatable} & \textbf{0.819} & \textbf{0.817} & 0.093 & \cellcolor{blue!15}0.177 & 0.062 \\
 & \multirow{1}{*}{\ctgan} & 0.816 & 0.813 & 0.097 & \cellcolor{blue!15}0.165 & 0.075 \\
 & \multirow{1}{*}{\tvae} & 0.815 & 0.813 & \cellcolor{blue!15}0.086 & \cellcolor{blue!15}0.171 & 0.062 \\
 & \multirow{1}{*}{\cart} & 0.816 & 0.813 & 0.097 & \cellcolor{blue!15}0.157 & 0.083 \\
 & \multirow{1}{*}{\smote} &  & & & & \\
\hline
\multirow{4}{*}{\classprotectedtable} & \multirow{1}{*}{\gaussiancopulatable} & \textbf{0.819} & \underline{0.816} & \cellcolor{blue!15}0.091 & \cellcolor{blue!15}0.178 & 0.060 \\
 & \multirow{1}{*}{\ctgan} & \underline{0.817} & 0.814 & 0.094 & \cellcolor{blue!15}0.167 & 0.072 \\
 & \multirow{1}{*}{\tvae} & 0.815 & 0.813 & \cellcolor{blue!15}0.086 & \cellcolor{blue!15}0.175 & 0.059 \\
 & \multirow{1}{*}{\cart} & 0.809 & 0.805 & 0.095 & \cellcolor{blue!15}\underline{0.154} & 0.080 \\
 & \multirow{1}{*}{\smote} &  & & & & \\
\hline
\multirow{4}{*}{\protectedonly} & \multirow{1}{*}{\gaussiancopulatable} & \textbf{0.819} & \underline{0.816} & \cellcolor{blue!15}0.090 & \cellcolor{blue!15}0.188 & 0.049 \\
 & \multirow{1}{*}{\ctgan} & \textbf{0.819} & \underline{0.816} & \cellcolor{blue!15}0.091 & \cellcolor{blue!15}0.188 & 0.049 \\
 & \multirow{1}{*}{\tvae} & \textbf{0.819} & \textbf{0.817} & \cellcolor{blue!15}0.089 & 0.189 & \cellcolor{blue!15}\underline{0.047} \\
 & \multirow{1}{*}{\cart} & 0.702 & 0.689 & \cellcolor{blue!15}\underline{0.081} & \cellcolor{blue!15}\textbf{0.090} & 0.071 \\
 & \multirow{1}{*}{\smote} &  & & & & \\
\hline
\multirow{4}{*}{\sameclass} & \multirow{1}{*}{\gaussiancopulatable} & 0.816 & 0.813 & \cellcolor{blue!15}0.091 & \cellcolor{blue!15}0.175 & 0.061 \\
 & \multirow{1}{*}{\ctgan} & 0.809 & 0.806 & \cellcolor{blue!15}0.090 & \cellcolor{blue!15}0.165 & 0.067 \\
 & \multirow{1}{*}{\tvae} & 0.809 & 0.806 & \cellcolor{blue!15}0.091 & \cellcolor{blue!15}0.168 & 0.064 \\
 & \multirow{1}{*}{\cart} & 0.807 & 0.804 & 0.099 & \cellcolor{blue!15}0.155 & 0.081 \\
 & \multirow{1}{*}{\smote} &  & & & & \\

\hline
\multicolumn{7}{c}{\cellcolor{black!15}\creditdataset\ dataset}\\
\hline

\samplingmethod & \training & \multicolumn{5}{c}{\metrics} \\
&  & \multicolumn{5}{c}{sex} \\
& & \acc & \rocauc & \eqoddtable & \statpartable & \eqopptable \\
\hline & \multirow{1}{*}{Real} & 0.812 & 0.651 & 0.034 & 0.023 & 0.021 \\
\hline
 & \multirow{1}{*}{\tabfairgan} & 0.784 & 0.597 & 0.042 & \cellcolor{blue!15}\underline{0.020} & 0.032 \\
\hline
\hline
\multirow{5}{*}{\classonly} & \multirow{1}{*}{\gaussiancopulatable} & 0.807 & \cellcolor{blue!15}0.662 & 0.045 & 0.026 & 0.028 \\
 & \multirow{1}{*}{\ctgan} & 0.807 & \cellcolor{blue!15}0.666 & 0.044 & 0.026 & 0.028 \\
 & \multirow{1}{*}{\tvae} & 0.804 & \cellcolor{blue!15}0.653 & 0.036 & 0.026 & \cellcolor{blue!15}\underline{0.019} \\
 & \multirow{1}{*}{\cart} & 0.757 & \cellcolor{blue!15}\underline{0.692} & 0.052 & 0.033 & 0.031 \\
 & \multirow{1}{*}{\smote} & 0.766 & \cellcolor{blue!15}0.666 & 0.037 & 0.023 & 0.024 \\
\hline
\multirow{5}{*}{\classprotectedtable} & \multirow{1}{*}{\gaussiancopulatable} & 0.806 & \cellcolor{blue!15}0.662 & 0.034 & 0.025 & \cellcolor{blue!15}\underline{0.019} \\
 & \multirow{1}{*}{\ctgan} & 0.809 & 0.644 & \cellcolor{blue!15}\underline{0.033} & \cellcolor{blue!15}\textbf{0.019} & 0.021 \\
 & \multirow{1}{*}{\tvae} & 0.806 & \cellcolor{blue!15}0.656 & 0.041 & \cellcolor{blue!15}0.021 & 0.029 \\
 & \multirow{1}{*}{\cart} & 0.760 & \cellcolor{blue!15}\textbf{0.693} & 0.054 & 0.032 & 0.034 \\
 & \multirow{1}{*}{\smote} & 0.769 & \cellcolor{blue!15}0.668 & 0.038 & \cellcolor{blue!15}0.021 & 0.026 \\
\hline
\multirow{5}{*}{\protectedonly} & \multirow{1}{*}{\gaussiancopulatable} & \cellcolor{blue!15}\underline{0.814} & 0.649 & 0.035 & 0.023 & 0.021 \\
 & \multirow{1}{*}{\ctgan} & \cellcolor{blue!15}\underline{0.814} & \cellcolor{blue!15}0.652 & 0.038 & 0.024 & 0.025 \\
 & \multirow{1}{*}{\tvae} & \cellcolor{blue!15}\textbf{0.815} & \cellcolor{blue!15}0.655 & 0.042 & 0.028 & 0.026 \\
 & \multirow{1}{*}{\cart} & \cellcolor{blue!15}0.813 & \cellcolor{blue!15}0.654 & 0.039 & 0.025 & 0.024 \\
 & \multirow{1}{*}{\smote} & \cellcolor{blue!15}0.813 & 0.647 & 0.040 & \cellcolor{blue!15}\textbf{0.019} & 0.027 \\
\hline
\multirow{5}{*}{\sameclass} & \multirow{1}{*}{\gaussiancopulatable} & \cellcolor{blue!15}0.813 & 0.649 & \cellcolor{blue!15}\textbf{0.032} & 0.023 & \cellcolor{blue!15}\textbf{0.018} \\
 & \multirow{1}{*}{\ctgan} & 0.810 & 0.632 & 0.034 & \cellcolor{blue!15}0.022 & \cellcolor{blue!15}0.020 \\
 & \multirow{1}{*}{\tvae} & \cellcolor{blue!15}\underline{0.814} & 0.650 & \cellcolor{blue!15}\underline{0.033} & \cellcolor{blue!15}0.021 & 0.021 \\
 & \multirow{1}{*}{\cart} & \cellcolor{blue!15}0.813 & 0.645 & 0.036 & \cellcolor{blue!15}0.022 & 0.022 \\
 & \multirow{1}{*}{\smote} & \cellcolor{blue!15}0.813 & 0.649 & 0.039 & 0.023 & 0.025 \\
\hline
\end{tabular}
\end{table*}

\begin{table*}[ht!]
\caption{Results for \adult\ dataset with \emph{race} and \emph{sex \& race} (intersectional) protected attributes.}
\label{tab:results_intersectional}
\begin{tabular}{l l c c c c c}
\hline
\multicolumn{7}{c}{\cellcolor{black!15}\adult\ dataset - race} \\
\hline
\samplingmethod & \training & \multicolumn{5}{c}{\metrics} \\
& & \acc & \rocauc & \eqoddtable & \statpartable & \eqopptable \\
\hline & \multirow{1}{*}{Real} & \textbf{0.868} & \underline{0.798} & 0.092 & 0.096 & 0.064 \\
\hline
\multirow{5}{*}{\classonly} & \multirow{1}{*}{\gaussiancopulatable} & 0.853 & 0.763 & \cellcolor{blue!15}0.073 & \cellcolor{blue!15}0.073 & \cellcolor{blue!15}0.049 \\
 & \multirow{1}{*}{\ctgan} & 0.850 & 0.786 & \cellcolor{blue!15}0.063 & \cellcolor{blue!15}0.084 & \cellcolor{blue!15}0.036 \\
 & \multirow{1}{*}{\tvae} & 0.846 & 0.771 & 0.104 & \cellcolor{blue!15}0.088 & 0.065 \\
 & \multirow{1}{*}{\cart} & 0.834 & \cellcolor{blue!15}\textbf{0.815} & \cellcolor{blue!15}0.081 & 0.113 & \cellcolor{blue!15}\underline{0.028} \\
 & \multirow{1}{*}{\smote} & 0.839 & 0.792 & 0.105 & 0.106 & \cellcolor{blue!15}0.052 \\
\hline
\multirow{5}{*}{\classprotectedtable} & \multirow{1}{*}{\gaussiancopulatable} & 0.849 & 0.758 & \cellcolor{blue!15}0.069 & \cellcolor{blue!15}0.072 & \cellcolor{blue!15}0.045 \\
 & \multirow{1}{*}{\ctgan} & 0.846 & 0.796 & \cellcolor{blue!15}0.057 & \cellcolor{blue!15}0.092 & \cellcolor{blue!15}\textbf{0.022} \\
 & \multirow{1}{*}{\tvae} & 0.847 & 0.766 & 0.097 & \cellcolor{blue!15}0.088 & \cellcolor{blue!15}0.060 \\
 & \multirow{1}{*}{\cart} & 0.821 & 0.790 & 0.100 & 0.107 & \cellcolor{blue!15}0.046 \\
 & \multirow{1}{*}{\smote} & 0.837 & 0.793 & 0.112 & 0.109 & \cellcolor{blue!15}0.057 \\
\hline
\multirow{5}{*}{\protectedonly} & \multirow{1}{*}{\gaussiancopulatable} & 0.849 & 0.761 & \cellcolor{blue!15}0.060 & \cellcolor{blue!15}0.075 & \cellcolor{blue!15}0.042 \\
 & \multirow{1}{*}{\ctgan} & 0.857 & 0.767 & \cellcolor{blue!15}0.068 & \cellcolor{blue!15}0.081 & \cellcolor{blue!15}0.045 \\
 & \multirow{1}{*}{\tvae} & 0.853 & 0.757 & \cellcolor{blue!15}0.066 & \cellcolor{blue!15}\underline{0.070} & \cellcolor{blue!15}0.045 \\
 & \multirow{1}{*}{\cart} & \underline{0.859} & 0.775 & \cellcolor{blue!15}0.067 & \cellcolor{blue!15}0.085 & \cellcolor{blue!15}0.043 \\
 & \multirow{1}{*}{\smote} & 0.856 & 0.757 & \cellcolor{blue!15}0.081 & \cellcolor{blue!15}0.080 & \cellcolor{blue!15}0.059 \\
\hline
\multirow{5}{*}{\sameclass} & \multirow{1}{*}{\gaussiancopulatable} & \underline{0.859} & 0.770 & \cellcolor{blue!15}\underline{0.056} & \cellcolor{blue!15}0.076 & \cellcolor{blue!15}0.034 \\
 & \multirow{1}{*}{\ctgan} & \underline{0.859} & 0.769 & \cellcolor{blue!15}\underline{0.056} & \cellcolor{blue!15}0.072 & \cellcolor{blue!15}0.036 \\
 & \multirow{1}{*}{\tvae} & 0.858 & 0.767 & \cellcolor{blue!15}0.070 & \cellcolor{blue!15}0.074 & \cellcolor{blue!15}0.048 \\
 & \multirow{1}{*}{\cart} & 0.856 & 0.767 & \cellcolor{blue!15}\textbf{0.050} & \cellcolor{blue!15}\textbf{0.069} & \cellcolor{blue!15}0.030 \\
 & \multirow{1}{*}{\smote} & 0.857 & 0.770 & \cellcolor{blue!15}0.058 & \cellcolor{blue!15}0.074 & \cellcolor{blue!15}0.034 \\

\hline
\multicolumn{7}{c}{\cellcolor{black!15}\adult\ dataset - sex \& race (intersectional)} \\
\hline

\samplingmethod & \training & \multicolumn{5}{c}{\metrics} \\
& & \acc & \rocauc & \eqoddtable & \statpartable & \eqopptable \\
\hline & \multirow{1}{*}{Real} & \textbf{0.867} & \underline{0.798} & 0.205 & 0.221 & 0.131 \\
\hline
\multirow{5}{*}{\classonly} & \multirow{1}{*}{\gaussiancopulatable} & 0.855 & 0.764 & 0.209 & \cellcolor{blue!15}0.186 & 0.146 \\
 & \multirow{1}{*}{\ctgan} & 0.849 & 0.774 & \cellcolor{blue!15}0.192 & \cellcolor{blue!15}0.192 & \cellcolor{blue!15}0.122 \\
 & \multirow{1}{*}{\tvae} & 0.848 & 0.773 & 0.242 & \cellcolor{blue!15}0.213 & 0.153 \\
 & \multirow{1}{*}{\cart} & 0.835 & \cellcolor{blue!15}\textbf{0.817} & 0.222 & 0.243 & \cellcolor{blue!15}\underline{0.114} \\
 & \multirow{1}{*}{\smote} & 0.840 & 0.790 & 0.247 & 0.230 & 0.143 \\
\hline
\multirow{5}{*}{\classprotectedtable} & \multirow{1}{*}{\gaussiancopulatable} & 0.850 & 0.757 & \cellcolor{blue!15}0.188 & \cellcolor{blue!15}0.182 & \cellcolor{blue!15}0.122 \\
 & \multirow{1}{*}{\ctgan} & 0.834 & \underline{0.798} & 0.211 & 0.235 & \cellcolor{blue!15}\textbf{0.110} \\
 & \multirow{1}{*}{\tvae} & 0.843 & 0.756 & 0.237 & \cellcolor{blue!15}0.189 & 0.163 \\
 & \multirow{1}{*}{\cart} & 0.825 & 0.797 & 0.252 & 0.243 & 0.133 \\
 & \multirow{1}{*}{\smote} & 0.839 & 0.797 & 0.253 & 0.252 & 0.135 \\
\hline
\multirow{5}{*}{\protectedonly} & \multirow{1}{*}{\gaussiancopulatable} & 0.841 & 0.752 & 0.219 & \cellcolor{blue!15}0.169 & 0.168 \\
 & \multirow{1}{*}{\ctgan} & 0.853 & 0.763 & 0.267 & \cellcolor{blue!15}0.213 & 0.188 \\
 & \multirow{1}{*}{\tvae} & 0.850 & 0.750 & 0.210 & \cellcolor{blue!15}0.180 & 0.147 \\
 & \multirow{1}{*}{\cart} & \underline{0.857} & 0.776 & 0.233 & \cellcolor{blue!15}0.210 & 0.162 \\
 & \multirow{1}{*}{\smote} & 0.852 & 0.742 & 0.263 & \cellcolor{blue!15}0.179 & 0.207 \\
\hline
\multirow{5}{*}{\sameclass} & \multirow{1}{*}{\gaussiancopulatable} & 0.854 & 0.757 & \cellcolor{blue!15}0.199 & \cellcolor{blue!15}0.180 & 0.139 \\
 & \multirow{1}{*}{\ctgan} & 0.854 & 0.764 & \cellcolor{blue!15}0.180 & \cellcolor{blue!15}0.178 & \cellcolor{blue!15}0.120 \\
 & \multirow{1}{*}{\tvae} & 0.855 & 0.764 & \cellcolor{blue!15}\underline{0.173} & \cellcolor{blue!15}0.182 & \cellcolor{blue!15}0.115 \\
 & \multirow{1}{*}{\cart} & 0.854 & 0.781 & \cellcolor{blue!15}\textbf{0.162} & \cellcolor{blue!15}\textbf{0.163} & \cellcolor{blue!15}0.118 \\
 & \multirow{1}{*}{\smote} & 0.853 & 0.768 & \cellcolor{blue!15}0.181 & \cellcolor{blue!15}\underline{0.167} & \cellcolor{blue!15}0.128 \\
\hline

\end{tabular}
\end{table*}

\section{Conclusion and discussion}
\label{sec:discussion}
Training ML models that take fairness and class imbalance into account is an open problem, with many applications in the real world, especially for tabular datasets. Most model-independent methods perform oversampling by building upon existing methods that generate synthetic samples via interpolation in the minority classes (SMOTE). This comparative study, considers several state-of-the-art generative approaches to synthesize tabular data in each minority class, to overcome bias. Results on four real-world tabular datasets indicate that the non-parametric \cart\ is the better-performing generative method while being the most computationally efficient. In future work, we would like to delve deeper into exploring the capabilities of \cart\ for generating synthetic data, particularly when optimized in the context of fairness.

% In future work, we plan to evaluate more complex scenarios of, for example, continuous protected attributes (e.g. age), or multi-class classification along with multi-fairness (multiple protected attributes). Such settings can potentially split the dataset into numerous scarce subgroups, posing significant challenges.

\clearpage

\section{Acknowledgements}
Our work is Funded by the Deutsche Forschungsgemeinsschaft (DFG, German Research Foundation ) - SFB1463 - 434502799. I further acknowledge the support by the European Union, Horizon Europe project MAMMOth under contract number 101070285.

\bibliographystyle{splncs04}
\bibliography{paper}

\end{document}